\documentclass[lettersize,journal]{IEEEtran}
\usepackage{amsmath,amsfonts}
\usepackage{algorithmic}
\usepackage{array}
\usepackage[caption=false,font=normalsize,labelfont=sf,textfont=sf]{subfig}
\usepackage{textcomp}
\usepackage{stfloats}
\usepackage{url}
\usepackage{bm}
\usepackage{verbatim}
\usepackage{booktabs}
\usepackage{multirow}
\usepackage{fontawesome5}
\usepackage{graphicx}
\usepackage{xcolor}
\usepackage{orcidlink}
\usepackage{hyperref}
\newcommand{\emoji}[1]{\raisebox{-0.25ex}{\includegraphics[height=1em]{#1}}}
\hypersetup{
colorlinks=True,
citecolor=blue,
linkcolor=blue,
hypertexnames=false,
}
\allowdisplaybreaks
\let\oldcite=\cite
\let\oldeqref=\eqref
\renewcommand{\cite}[1]{\textcolor{blue}{\oldcite{#1}}}
\renewcommand{\eqref}[1]{\textcolor{blue}{\oldeqref{#1}}}
\def\BibTeX{{\rm B\kern-.05em{\sc i\kern-.025em b}\kern-.08em
    T\kern-.1667em\lower.7ex\hbox{E}\kern-.125emX}}
\usepackage{balance}
\begin{document}
\title{ReSemAct: Advancing Fine-Grained Robotic Manipulation via Semantic Structuring and Affordance Refinement}
\author{
Chenyu Su,
Weiwei Shang,
Chen Qian,
Fei Zhang,
and Shuang Cong
\thanks{The authors are with the Department of Automation, University of Science and Technology of China, Hefei 230027, China (e-mail: suchenyu@mail.ustc.edu.cn; wwshang@ustc.edu.cn; qian\_chen@ustc.edu.cn; zfei@ustc.edu.cn; scong@ustc.edu.cn). Corresponding author: Weiwei Shang.}
}
\IEEEaftertitletext{
\vspace{-1.5\baselineskip}
\begin{center}

\faGithub~\textbf{Code:} \href{https://github.com/scy-v/ReSemAct}{\textbf{\texttt{github.com/scy-v/ReSemAct}}}\hspace{1em}
\emoji{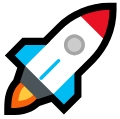}~\textbf{Homepage:} \href{https://resemact.github.io}{\textbf{\texttt{ReSemAct.github.io}}}
\end{center}
\vspace{1.5\baselineskip}
}

\maketitle

\begin{abstract}
Fine-grained robotic manipulation requires grounding natural language into appropriate affordance targets. However, most existing methods driven by foundation models often compress rich semantics into oversimplified affordances, preventing exploitation of implicit semantic information. 
To address these challenges, we present ReSemAct, a novel unified manipulation framework that introduces Semantic Structuring and Affordance Refinement (SSAR), powered by the automated synergistic reasoning between Multimodal Large Language Models (MLLMs) and Vision Foundation Models (VFMs). Specifically, the Semantic Structuring module derives a unified semantic affordance description from natural language and RGB observations, organizing affordance regions, implicit functional intent, and coarse affordance anchors into a structured representation for downstream refinement. Building upon this specification, the Affordance Refinement strategy instantiates two complementary flows that separately specialize geometry and position, yielding fine-grained affordance targets. These refined targets are then encoded as real-time joint-space optimization objectives, enabling reactive and robust manipulation in dynamic environments.
Extensive simulation and real-world experiments are conducted in semantically rich household and sparse chemical lab environments. The results demonstrate that ReSemAct performs diverse tasks under zero-shot conditions, showcasing the robustness of SSAR with foundation models in fine-grained manipulation.

\end{abstract}


\section{Introduction}
\IEEEPARstart{W}{ith} the advancement of robotic autonomy in unstructured environments, the growing diversity of semantic descriptions and the complexity of visual modeling have become critical challenges limiting system generalization and robustness \cite{walke2023bridgedata}. Recent advances in cross-modal reasoning powered by Multimodal Large Language Models (MLLMs) \cite{achiam2023gpt-4} and Vision Foundation Models (VFMs) \cite{zhao2023fastsam} have significantly improved semantic grounding, which is crucial for constructing accurate affordance targets. For instance, tweezers-based grasping entails localizing the coarse semantic grasp region and reasoning about the geometric relationships between the tweezer tips and the grasp point. Based on these inferred relations, affordance targets can be constructed and parsed in real-time by the constraint optimizer, thereby enabling closed-loop control and fine-grained manipulation.

However, existing approaches typically map semantic information into coarse-grained affordance representations, lacking an optimizable intermediate structure that enables fine-grained semantic grounding \cite{ju2025roboabc,li2025affgrasp}. Although Multimodal Large Language Models (MLLMs) have shown potential in cross-modal reasoning, their geometric inference and spatial localization capabilities remain insufficient for directly deriving fine-grained targets from natural language \cite{li2024manipllm,liu2024moka}. Language-conditioned functional grasping \cite{tang2025affordgrasp,tang2025foundationgrasp} provides partial associations between semantic intent and action execution, yet its expressiveness is confined to part-level granularity, limiting complex manipulation. Prior efforts have attempted to integrate MLLMs with VFMs to enhance language–vision semantic alignment \cite{yang2023setofmark,yang2025magma} and support the automated construction of affordance targets, but these targets largely remain at coarse localization, lacking semantic-guided refinement \cite{huang2023voxposer,huang2024rekep}. This highlights the urgent need for fine-grained semantic grounding to enable robotic manipulation.
Inspired by these observations, we note that task-oriented grasping often implicitly encodes functional structures relevant to the scenario, while foundation models exhibit flexible semantic–visual conditioning. Based on this, we raise two key questions: 1) how to represent affordances with semantically flexible descriptions to support fine-grained grounding, and 2) how to develop a closed-loop framework that enables autonomous semantic grounding and execution in dynamic environments.

To address the first challenge, we propose Semantic Structuring and Affordance Refinement (SSAR). Specifically, the Semantic Structuring module derives a unified semantic affordance description by leveraging the synergistic reasoning of MLLMs and VFMs to extract affordance mask regions, implicit functional intent, and coarse affordance targets through multimodal reasoning, thereby forming structured semantic representation. Building upon this, the Affordance Refinement module instantiates geometric and positional flows to analyze the structured semantics affordance automatically, progressively refining the coarse affordance targets. Subsequently, the refined targets are encoded as optimization objectives within a Model Predictive Path Integral (MPPI) control \cite{zhang2024mppi} running in real time on the Isaac Gym platform \cite{makoviychuk2021isaacgym}, with joint velocity optimization performed at 15 Hz.

To address the second challenge, we develop a Task and Motion Planning (TAMP) framework. This framework extends the hierarchical recursive prompting structure of Liang et al. \cite{liang2023cap} and is redeveloped under MLLM‑based reasoning, enabling the autonomous operation of SSAR and closed‑loop robotic execution. The framework automatically extracts and parses semantic information, supports multi‑stage task decomposition, and generates the necessary precondition checks, postcondition evaluations, and optimizable cost functions for each subtask. Furthermore, we propose a subtask‑level backtracking strategy that leverages precondition feasibility to trigger cross‑stage replanning, thereby achieving behavior recovery and reactive control under dynamic perturbations.

The contributions of this work are summarized as follows:
\begin{enumerate}
    \item We propose Semantic Structuring and Affordance Refinement (SSAR), a new method that introduces the semantic affordance description through multimodal reasoning and the Affordance refinement flow to enable fine-grained semantic grounding.
    \item Closed-loop control for diverse tasks with fine-grained semantic grounding is achieved with the newly developed MLLM-driven TAMP framework, which performs autonomous multi-stage task decomposition with condition reasoning and cost optimization.
    \item Through extensive simulations and real-world experiments in semantically rich household and sparse chemical lab environments, we successfully validate ReSemAct's semantic grounding capabilities, demonstrating strong generalization.
\end{enumerate}

The remainder of this paper is organized as follows: In Section \ref{Section II}, we review related works including foundation models and robotic manipulation. In Section \ref{Section III}, we present the proposed system architecture in detail. In Section \ref{Section IV} we demonstrate the performance of our method in semantically diverse environments, chemical and household, through simulation and real-world experiments. In Section \ref{Section V}, we discuss the advantage, limitations and future work. Finally,  we conclude this work in Section \ref{Section VI}.

\section{Related Works}
\label{Section II}
\subsection{Semantic Grounding with Foundation Models}

Semantic grounding refers to mapping semantic features from natural language to image regions and spatial entities. Prior VFMs focused on image perception, leveraging self-supervised learning \cite{chen2020cl,he2020momentumconstrast} and architectures including CNNs \cite{lecun2002gradient,krizhevsky2017imagenet} and Vision Transformers \cite{dosovitskiy2020image}, which produce visual encodings for downstream tasks like object detection \cite{girshick2014rich,redmon2016yolo}, segmentation \cite{long2015fullyconv,bad2017segnet}, and feature extraction \cite{touvron2021img}. Furthermore, pixel-level parsing was enhanced through latent feature encoding and embedded visual prompts \cite{chen2022focalclick,zou2023samonce}. Subsequently, Vision-Language Models (VLMs) perform multimodal semantic grounding through visual-linguistic representations, achieving progress in image-text retrieval \cite{radford2021learntrans} and open-vocabulary object detection \cite{liu2024groundingdino}, but still struggle with fine-grained semantic grounding. Building on this, researchers introduced Large Language Models (LLMs) \cite{touvron2023llama} into vision-language systems, leading to MLLMs \cite{liu2023visualins} that incorporate visual features as context for multimodal reasoning. Despite this, MLLMs remain dependent on front-end visual encoders, limiting spatial precision and semantic understanding. Recent studies have explored the synergy between VFMs and MLLMs by embedding visual prompts for semantic alignment \cite{yang2023setofmark}, but fine-grained semantic grounding remains a critical limitation. In contrast to the above methods, our work focuses on a structured semantic representation together with further refinement flows, framing fine-grained semantic grounding as an explicit and progressively optimized grounding strategy.

\subsection{Affordance-Based Robotic Manipulation}

Affordance are essential for robotic manipulation, bridging high-level semantics and low-level actions to ensure task feasibility. 
Conventional methods construct affordance through explicit geometric modeling of object shapes, contact dynamics, and environmental structure \cite{karlbling2013tamp,kopicki2011manip} to enable motion planning. 
While physically interpretable and effective in structured scenarios, their dependence on accurate modeling constrains their flexibility in complex settings.
Recently, perception-driven learning has been used to infer potential manipulation regions from visual observations, including affordance recognition \cite{chu2019afford}, grasp pose detection \cite{fang2023anygrasp,pav2025grasp}, target pose estimation \cite{zheng2023hs-pose}, and semantic keypoint generation \cite{schmidt2016descriptor}. These approaches have  further extended to multimodal manipulation frameworks that incorporate foundation model reasoning \cite{tang2023graspgpt,li2025graspdex}. While capable of adapting to object deformations and semantic changes, they still require manual data collection and struggle with fine-grained semantic parsing and precise modeling of complex geometric relationships. Current approaches integrate language and vision models to enable zero-shot affordance region perception \cite{huang2024copa,huang2024rekep}. 
Nevertheless, existing methods predominantly employ coarse affordance representations. By contrast, our work emphasizes semantic affordance descriptions that capture implicit intent, affordance regions, and affordance anchors, enabling the construction of semantically grounded affordance targets.

\subsection{Task and Motion Planning}	

Task and Motion Planning (TAMP) provides a pivotal framework connecting high-level task reasoning with low-level motion execution. Conventional approaches often leverage formal languages such as PDDL and HTN to model symbolic tasks, solving task sequences and motion trajectories through logic-geometric and mixed-integer programming \cite{dantam2016tamp,garrett2018tamp}. Although  offering strong interpretability, these methods rely on manually designed task models and action primitives, limiting their adaptability in open environments. With the advancement of LLMs, recent works have explored zero-shot task planning based on predefined motion primitives \cite{singh2023progprompt}. However, these approaches lack effective modeling of geometric constraints and environmental dynamics. Building on these developments, recent research has explored integrating LLMs with motion planning to enable high-level task decomposition and continuous action generation without predefined primitives \cite{huang2023voxposer}. Although such methods enhance the flexibility of TAMP, they often depend on online optimization in task space, posing limitations for real-time execution. Our work implements an automated TAMP framework driven by MLLMs that couples high-level task reasoning with real-time joint-space constraint optimization, where subtask-level constraints and objectives are automatically generated to support closed-loop execution.

\begin{figure*}[!t]
\centering
\includegraphics[width=0.85\textwidth]{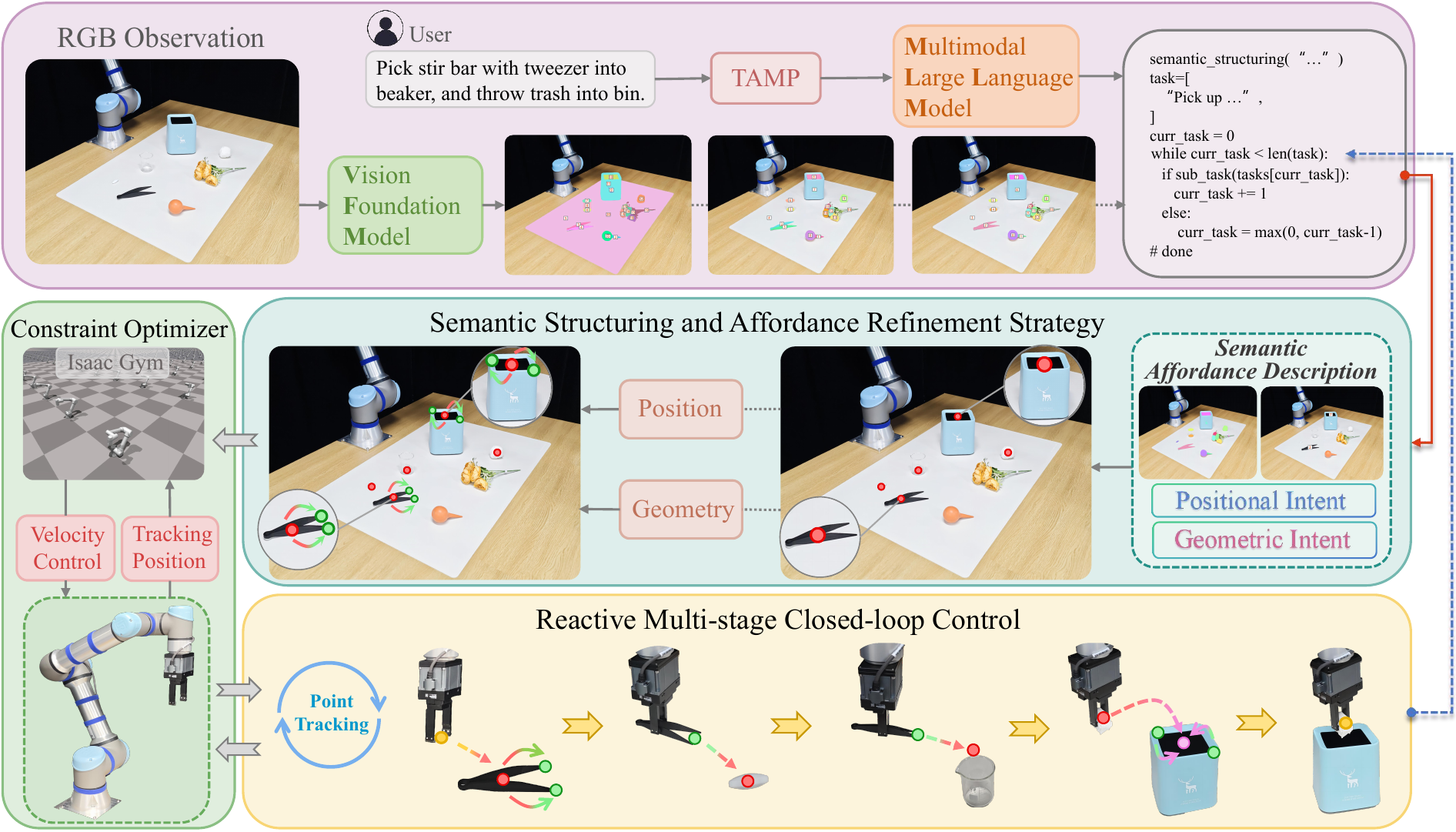}
\caption{Overall Framework. Given natural language instructions and RGB-D observations, the synergistic reasoning between VFM and MLLM facilitates the semantic structuring representation. Within the MLLM-driven TAMP framework, the Semantic Structuring and Affordance Refinement Strategy performs geometric and positional flow optimization over the Semantic Affordance Description, thereby generating the refined affordance targets. These targets are then encoded as cost functions and solved in real-time by a closed-loop MPPI-based optimizer in Isaac Gym, enabling joint-space control with point tracking.}
\label{fig_1}
\end{figure*}

\section{Method}
\label{Section III}
In this section, we first formulate ReSemAct as a real-time constrained optimization problem with semantic grounding. Subsequently, we introduce Semantic Structuring and Affordance Refinement, which enables fine-grained semantic grounding through multimodal reasoning. Thereafter, we derive a real-time closed-loop control strategy for solving the optimization problem. Finally, we integrate MLLM-driven reasoning to automate TAMP, enabling closed-loop execution.

\subsection{ReSemAct Problem Formulation}

As shown in Fig.~\ref{fig_1}, given a free-form natural language instruction $\mathcal{L}=\{L_1,L_2,\ldots,L_n\}$ from the user, $\mathcal{L}$ typically exhibits significant long-term dependencies and complex semantic reasoning. Simultaneous generation and optimization of affordance targets compromises reasoning accuracy and leads to overly coarse granularity.
 Therefore, ReSemAct employs a hierarchical modeling strategy by decomposing the task into sequential subtasks and progressively performs affordance reasoning and optimization for each $L_i$, $i\in\{1,2,\ldots,n\}$.

Specifically, for the $i$-th subtask instruction $L_i$, the instantiation of the semantic grounding is defined as

\begin{equation}
\label{eq1}
f: (L_i, O_i)\rightarrow\mathcal{C}_i^{\text{init}}
\end{equation}

\noindent where $O_i$ denotes the RGB-D sensor observations, $\mathcal{C}_i^{\text{init}}\in SE(3)$ denotes the initial set of affordace targets mapped by $f$. 
In order to capture uncertainties in dynamic environments, we introduce external dynamic evolution and disturbances $\mathcal{E}_i$. Additionally, we define $\bm{T}_e(t)\in SE(3)$ as the end-effector pose at time $t$, and $S$ as the known kinematic and physical model. At each stage $i$, ReSemAct addresses task execution by resolving the constrained optimization problem in real-time:
\begin{align}
\label{eq2}
\operatorname*{argmin}_{\bm{v}(t)} \quad 
& \mathcal{J}\left(\bm{T}_e(t), \mathcal{C}_i\right) \\
\text{subject to} \quad 
& \mathcal{C}_i^{\text{init}} = f(L_i, O_i), \notag \\
& \mathcal{C}_i = g(\mathcal{C}_i^{\text{init}}, \mathcal{E}_i), \notag \\
& \epsilon_{\text{pre}}\!\left(\bm{T}_e(t), \mathcal{C}_i\right) \le \varepsilon_{\text{pre}}, \notag \\
& \epsilon_{\text{post}}\!\left(\bm{T}_e(t), \mathcal{C}_i\right) \le \varepsilon_{\text{post}}, \notag \\
& \bm{v}(t),\ \bm{T}_e(t) \in \text{Feasible}(S) \notag
\end{align}

\noindent where $\mathcal{J}\left(\bm{T}_e(t), \mathcal{C}_i\right)$ is the real-time optimized cost function, $g\left(\cdot\right)$ denotes the instantiation mapping that updates affordance targets based on external disturbances, $\epsilon_{\text{pre}}\left(\bm{T}_e(t), \mathcal{C}_i\right)$ and $\epsilon_{\text{post}}\left(\bm{T}_e(t), \mathcal{C}_i\right)$ represent the precondition and postcondition constraints for subtask execution, with $\varepsilon_{\text{pre}}$ and $\varepsilon_{\text{post}}$ specifying their tolerance thresholds.
The set $\text{Feasible}\left(S\right)$ enforces consistency with the robot kinematic and physical model, $\bm{v}(t)\in \mathbb R^n$ is the joint space velocity, serving as the decision variable, with $n$ denoting the degrees of freedom.

Building on the above configurations, 
ReSemAct decomposes long-horizon manipulation tasks into multi-stage real-time constrained optimization problems, achieving semantic grounding and closed-loop dynamic control for each sub-task. 
Subsequently, the processed decision variable $\bm{v}(t)$ directly serves as the joint velocity command, enabling the robot to perform tasks effectively in dynamic environments.

\begin{figure*}[!t]
\centering
\includegraphics[width=0.85\textwidth]{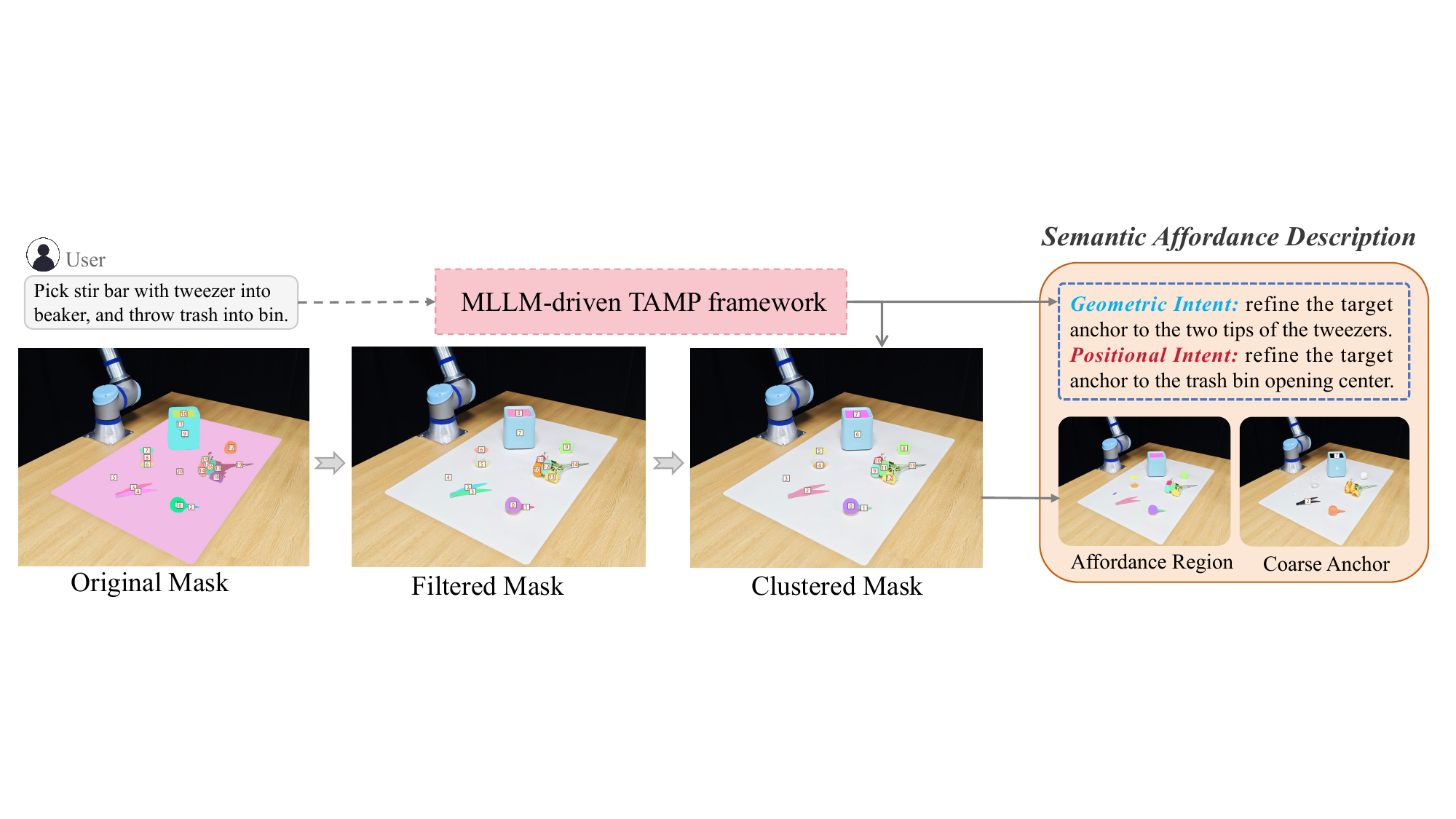}
\caption{Semantic Structuring Representation. By extracting the implicit intent of instructions, the affordance regions from scene masks, and the coarse-grained affordance anchors of objects, the semantic affordance description is constructed within the MLLM-driven TAMP framework.
}
\label{fig_2}
\end{figure*}

\subsection{Semantic Structuring and Affordance Refinement}

To instantiate $f:(L_i, O_i)\rightarrow\mathcal{C}_i^{\text{init}}$, we leverages multimodal reasoning to produce the semantic affordance description, which is transformed via geometric and positional refinement flows to obtain fine-grained semantic grounding.

\subsubsection{Semantic Structuring Representation}

As illustrated in Fig.~\ref{fig_2}, we define a semantic affordance description to construct an optimizable and interpretable semantic structure for downstream target refinement: the affordance region indicates the interactable space, the implicit functional intent specifies the fine-grained requirements, and the coarse affordance anchor provides the initial target for optimization.

\textbf{Implicit Semantic Intent.} To effectively leverage the implicit intent embedded in the given subtask instruction $L_i$, we extract its geometric and positional cues and reformulate them as an implicit functional intent $S_{\text{intent}}$ through recursive reasoning within the MLLM-driven TAMP framework.

\textbf{Affordance Region.} Given RGB-D observation $O_i\in\mathbb{R}^{H\times W\times4}$, we apply the pretrained image segmentation model \cite{zhao2023fastsam} to produce the initial set of masks $M =\{m_1, m_2, \ldots, m_g\}$, where $m_j\in{\{0,\ 1\}}^{H\times W}$ and $\forall j\in\{1,\ldots,g\}$. Here, $H$ and $W$ denote the height and width of the image, respectively. To extract the affordance regions $M_{\text{region}}$ of objects, the following two steps are designed:

\textbf{Step 1: Mask Filtering.}  A dual-filtering pipeline is applied to promote valid and distinguishable masks, resulting in the refined mask set $M_{\text{filtered}}$.

a) Area-based Filtering: we retain masks within the valid area range and define the mask set as
\begin{equation}
\label{eq3}
M_{\text{area}}=\{m_j\in M\ |\ \alpha \cdot A_{\text{img}}\le \text{Area}(m_j)\le\beta \cdot A_{\text{img}}\}
\end{equation}

\noindent where $A_{\text{img}}$ denotes the image area, $\alpha$ and $\beta$ are thresholds.

b) Structural Independence Filtering: we omit any mask that contains more than a specified number of other valid masks, and define the subset $M_{\text{contain}}$ as
\begin{equation}
\label{eq4}
M_{\text{contain}}=\{m_j\in M | N_{\text{sub}}\left(m_j\right)<3\}
\end{equation}

\noindent in which $N_{\text{sub}}\left(m_j\right)$ denotes the count of masks contained within mask $m_j$, defined as
\begin{equation}
\label{eq5}
N_{\text{sub}}\left(m_j\right)=|\{m_k\in M | m_k\subset m_j\}|
\end{equation}

Finally, The resulting effective mask set after the dual-filtering step is defined as
\begin{equation}
\label{eq6}
M_{\text{filtered}}=M_{\text{area}}\cap M_{\text{contain}}
\end{equation}

\textbf{Step 2: Visual Consistency Clustering.} Density-based clustering (DBSCAN) is performed on the filtered mask set to further reduce visual ambiguity and redundancy, with masks of identical semantic labels merged within each cluster. The clustered mask set is denoted as
\begin{equation}
\label{eq7}
M_{\text{cluster}}=\text{DBSCAN}(M_{\text{filtered}})	
\end{equation}

\noindent where $M_{\text{cluster}}=\{{\widetilde{m}}_1, {\widetilde{m}}_2, \ldots, {\widetilde{m}}_e\}$, and $e$ is the cardinality. 

\textbf{Coarse Affordance Anchor.} Since affordance targets requires reasoning within affordance regions while remaining consistent with implicit semantic intent, we introduce a coarse affordance anchor to instantiate an optimizable starting point for downstream refinement. We define the centroid coordinate $\bm{c}_j$ for each clustered mask ${\widetilde{m}}_j\in M_{\text{cluster}}$ as
\begin{equation}
\label{eq8}
\bm{c}_j = \frac{1}{|\widetilde{m}_j|} \sum_{(x, y) \in \widetilde{m}_j} 
\begin{bmatrix}
x \\ y
\end{bmatrix}
\end{equation}

\noindent where \((x, y)\) denotes the 2D coordinates of the pixels in the mask \(\widetilde{m}_j\), and \(|\widetilde{m}_j|\) is the number of pixels in \(\widetilde{m}_j\). Each centroid in the coordinate set $C=\{\bm{c}_{1}, \bm{c}_{2}, \ldots, \bm{c}_{q}\}$ is sequentially assigned a numeric label from the label set $\mathcal{Z}=\{0, 1, \ldots, q-1\}$ to enhance semantic grounding of MLLM. These labels are overlaid onto the original RGB image $I_{\text{rgb}}\in\mathbb{R}^{H\times W\times3}$ to form an explicit visual prompt. Together with the subtask instruction $L_i$, the resulting multimodel input is fed into the MLLM, which leverages semantic grounding to extract the labels relevant to the instruction, thereby obtaining the affordance anchor $C_{\text{anchor}} \subseteq C$. The $M_{\text{region}} \subseteq M_{\text{cluster}}$ associated with $C_{\text{anchor}}$ is constructed as the affordance region. Finally, the extracted labels are projected into 3D space via the depth information $I_{\text{depth}}\in\mathbb{R}^{H\times W}$, completing the instantiation of $C_{\text{anchor}}$. Based on the above formulation, the semantic affordance description is represented as $\mathcal{A}=\{S_{\text{intent}}, M_{\text{region}}, C_{\text{anchor}}\}$.

\subsubsection{Affordance Refinement Strategy}

\begin{figure*}[!t]
\centering
\includegraphics[width=0.85\textwidth]{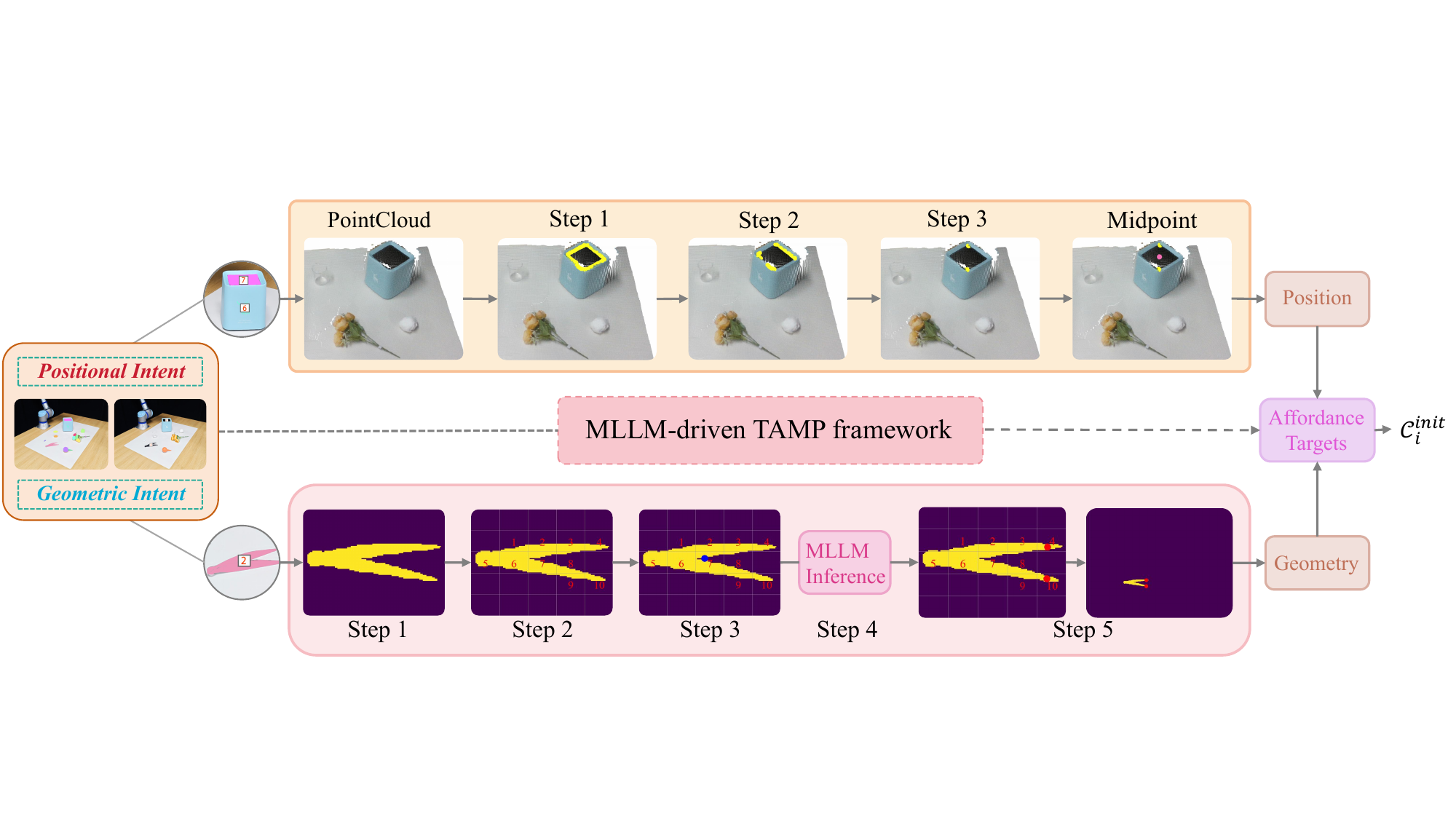}
\caption{Affordance Refinement Strategy.  Within the MLLM-driven TAMP framework, this module encompasses two strategies: geometric and positional refinement. The geometric refinement adjusts the tweezers’ grasp point from the center to the two tips, thereby introducing detailed geometric priors. The positional refinement localizes the trash bin’s placement to the midpoint between symmetrical centers, thereby enhancing the spatial precision.}
\label{fig_3}
\end{figure*}

As shown in Fig.~\ref{fig_3}, to effectively leverage the semantic affordance description $\mathcal{A}$ and further refine the coarse affordance anchor $C_{\text{anchor}}$, we propose an Affordance Refinement Strategy driven by the MLLM. This strategy automatically parses $\mathcal{A}$ and selects between two refinement flows: a geometric refinement flow, which focuses on object-level geometric structure, and a positional refinement flow, which emphasizes object opening localization. 
In addition, the instantiated $C_{\text{anchor}}$ will be directly used as the affordance targets $\mathcal{C}_i^{\text{init}}$ when the implicit intent cannot be resolved by the MLLM.

\textbf{Geometric Refinement Flow:} Given the semantic affordance description $\mathcal{A}=\{S_{\text{intent}}, M_{\text{region}}, C_{\text{anchor}}\}$, the geometric refinement flow is applied to enhance fine-grained semantic grounding and proceeds through the following five steps.

\textbf{Step 1: Mask Normalization.} Proportional discrepancies among different masks ${\widetilde{m}}_j \in M_{\text{region}}$ are mitigated by cropping the effective region of each mask and uniformly scaling them, thereby embedding all masks consistently onto the fixed-resolution standard canvas. Specifically, we first extract coordinates of the foreground pixels in ${\widetilde{m}}_j$ as
\begin{equation}
\label{eq9}
   \mathcal{S}_j=\{(x, y)\ |\ {\widetilde{m}}_j\left(x,y\right)=1\}
\end{equation}
\noindent where $\mathcal{S}_j$ denotes the set of foreground pixel coordinates. Subsequently, we determine the effective region of ${\widetilde{m}}_j$ by computing the boundary parameters of $\mathcal{S}_j$ as
\begin{align}
\nonumber
    t_j=\operatorname*{min}_{(x,y)\in\mathcal{S}_j}{y}, \quad b_j=\operatorname*{max}_{(x,y)\in\mathcal{S}_j}{y} \\
    \label{eq10}
    l_j=\operatorname*{min}_{(x,y)\in\mathcal{S}_j}{x},  \quad
    r_j=\operatorname*{max}_{(x,y)\in\mathcal{S}_j}{x} 
\end{align}

\noindent here $t_j$, $b_j$, $l_j$ and $r_j$ denote the top, bottom, left, and right boundaries, respectively. The mask ${\widetilde{m}}_j$ is then cropped according to these boundaries and resized to a fixed resolution $H_r\times W_r$. The cropped mask ${\widetilde{m}}_j^c={\widetilde{m}}_j\left(t_j:b_j,l_j:r_j\right)$ has dimensions $W_c=r_j-l_j+1$ in width and $H_c=b_j-t_j+1$ in height, where the scaling factors are defined as
\begin{equation}
\label{eq11}
\alpha_x=\frac{W_r}{W_c},\ \ \ \alpha_y=\frac{H_r}{H_c}
\end{equation}

\noindent in which $\alpha_x$ and $\alpha_y$ denote the scaling factors in the horizontal and vertical directions, respectively. Subsequently, we apply nearest-neighbor interpolation to resize the mask to the target dimensions as
\begin{equation}
\label{eq12}
{\widetilde{m}}_j^r=\mathcal{R}({\widetilde{m}}_j^c,\ \alpha_x,\alpha_y)
\end{equation}

\noindent where $\mathcal{R}$ denotes nearest-neighbor interpolation, and ${\widetilde{m}}_j^r$ is the resized mask. To maintain clear mask boundaries and prevent excessive fitting to image edges, the resizing resolution is usually smaller than the target canvas resolution. Consequently, the resized mask is further aligned onto the standardized canvas. Given the target resolution of $H_t\times W_t$, the target mask ${\widetilde{m}}_j^t$ is defined as
\begin{equation}
\label{eq13}
{\widetilde{m}}_j^t={\widetilde{m}}_j^r\left(y-\Delta_y,x-\Delta_x\right)
\end{equation}

\noindent where $\Delta_x=W_t-W_r/2$ and $\Delta_y=H_t-H_r/2$ denote the horizontal and vertical offsets, respectively, which ensure the mask remains centered within the image.

\textbf{Step 2: Grid Construction and Label Embedding.} In order to improve semantic grounding of geometric structures, grid construction and label embedding are performed on the target mask. Specifically, the $j$-th mask ${\widetilde{m}}_j^t$ is divided into $G_h\times G_w$ grids, where the width $\omega_c$ and height $h_c$ of each grid are determined by the target resolution $H_t\times W_t$ as follows:
\begin{equation}
\label{eq14}
\omega_c=\frac{W_t}{G_w},\ \ \ h_c=\frac{H_t}{G_h}
\end{equation}

For the arbitrary grid at the $k$-th row and $s$-th column, where $k\in\{0, \ldots, G_h-1\}$, $s\in\{0, \ldots, G_w-1\}$, the coordinate region covered by the grid is expressed as
\begin{equation}
\label{eq15}
\mathcal{G}_{k,s}^j = \left\{ (x, y)\ \middle|\
\begin{aligned}
&k \cdot h_c \le y < (k+1) \cdot h_c,\ \\
&s \cdot w_c \le x < (s+1) \cdot w_c
\end{aligned}
\right\}
\end{equation}

The foreground density $\rho_{k,s}^j$ for each $\mathcal{G}_{k,s}^j$ is determined by
\begin{equation}
\label{eq16}
\rho_{k,s}^j=\frac{\mathcal{P}_{k,s}^j}{w_c\cdot h_c} 
\end{equation}

\noindent where $\mathcal{P}_{k,s}^j = \sum_{(x,y) \in \mathcal{G}_{k,s}} \left[ \tilde{m}_j^t(x, y) = 1 \right]$ denotes the set of foreground pixels within $\mathcal{G}_{k,s}^j$, and $|\mathcal{P}_{k,s}^j|$ is the number of pixels. When the foreground density $\rho_{k,s}^j$ surpasses the assignment threshold $\tau$, the semantic label $\ell_{k,s}^j\in\left\{0,\ 1,\ldots,N_{\text{valid}}-1\right\}$ is assigned to the grid. We define the mapping $\mathcal{U}:\ \mathcal{G}_{k,s}^j\rightarrow\ell_{k,s}^j$  to associate each grid with the semantic label, where $N_{\text{valid}}$ indicates the total number of grids satisfying the threshold condition. The centroid within the grid are then calculated as
\begin{equation}
\label{eq17}
\bm{c}_j^{k,s}=\frac{1}{|\mathcal{P}_{k,s}^j|}\sum_{(x,y)\in\mathcal{P}_{k,s}^j}{\begin{bmatrix}
x \\
y
\end{bmatrix}}
\end{equation}

\textbf{Step 3: Centroid mapping and label association.} The course anchor $\bm{c}_j\in C_{\text{anchor}}$ are mapped onto the target mask and associated with the grid semantic labels $\ell_{k,s}^j$ , which are overlaid on the target mask as textual prompts to enhance the fine-grained geometric understanding of MLLM. Subsequently, we map the anchor $\bm{c}_j = [x_j \; y_j]^\top$ into the the target mask as follows:
\begin{equation}
\label{eq18}
{\bar{x}}_j=\frac{x_j-l_j}{\alpha_x}+\Delta_x,\ {\bar{y}}_j=\frac{y_j-t_j}{\alpha_y}+\Delta_y
\end{equation}

\noindent where $\bar{\bm{c}}_j=[\bar{x}_j \; \bar{y}_j]^\top$ denotes the anchor in the target mask. To establish the correspondence with grid labels, $\bar{\bm{c}}_j$ is assigned to its corresponding grid $\mathcal{G}_{k,s}^j$ by Equation \eqref{eq15}, and then projected to the grid centroid $\bm{c}_j^{k,s}$ following Equation \eqref{eq17}, thereby associating it to the semantic label $\ell_{k,s}^j$.

\textbf{Step 4: Fine-grained Geometric Reasoning by MLLM.} To generate new labels ${\hat{\ell}}_{k,s}^j$ based on the geometric structure, we feed the original RGB image $I_{\text{rgb}}$, the original mask ${\widetilde{m}}_j$, and the target mask ${\widetilde{m}}_j^t$ as visual prompts into the MLLM. Subsequently, the numerical label $z_j \in \mathcal{Z}$, the semantic label $\ell_{k,s}^j$, and the implicit intent $s_{\text{g}} \in S_{\text{intent}}$ are provided as textual prompts to the MLLM for reasoning. 

\textbf{Step 5: Refined Anchors as Affordance Targets.} Through the instantiated mapping $\mathcal{U}:\mathcal{G}_{k,s}^j \rightarrow \ell_{k,s}^j$, the newly generated labels ${\hat{\ell}}_{k,s}^j$ are inversely mapped to their corresponding grids $\mathcal{G}_{k,s}^j$, thereby constructing the refined anchor. Subsequently, the associated grid centroids $\bm{c}_j^{k,s}$ are projected to the original mask coordinate as $\bm{c}_j$. Finally, the 2D coordinates are transformed into 3D space under the depth information $I_{\text{depth}} \in \mathbb{R}^{H \times W}$, forming the affordance targets $\mathcal{C}_i^{\text{init}}$.

\textbf{Positional Refinement Flow:} To mitigate positional discrepancies in semantic grounding of objects characterized by distinct boundary features, including open-top structures, we propose a three-step refinement method that exploits edge points and affordance regions to optimize $\mathcal{A}=\{S_{\text{intent}}, M_{\text{region}}, C_{\text{anchor}}\}$.
By integrating density peak estimation with symmetry analysis, this approach effectively mitigates MLLM limitations in position modeling, enhancing the practical feasibility of semantic grounding.

\textbf{Step 1: Extraction of 3D Coordinates Along Mask Edges.} The edge point set $\mathcal{M}_j$ associated with mask ${\widetilde{m}}_j \in M_{\text{region}}$ is defined for extracting edge points of the object as follows:
\begin{equation}
\label{eq19}
\mathcal{M}_j=\{\left(x,y\right)\ |\ {\widetilde{m}}_j(x,y)=1\}  
\end{equation}

\noindent subject to:
\begin{equation}
\exists(x\prime,y\prime)\in\mathcal{N}_4\left(x,y\right) \quad \text{s.t.}{\ \widetilde{m}}_j(x\prime,y\prime)=0 \notag
\end{equation}

\noindent where $\mathcal{N}_4\left(x,y\right)$ denotes the 4-neighborhood of pixel coordinate $(x, y)$, defined as
\begin{equation}
\label{eq20}
\mathcal{N}_4\left(x,y\right)=\{\left(x\pm1,y\right),\left(x,y\pm1\right)\}
\end{equation}

Subsequently, with the depth information $I_{\text{depth}}\in\mathbb{R}^{H\times W}$, the set of 3D edge points $\mathcal{P}_j$ is mapped derived as follows:
\begin{equation}
\label{eq21}
\mathcal{P}_j=\{\mathcal{F}_{\text{3D}}\left(x,y,I_{\text{depth}}\right) | 
\,\forall(x,y)\in\mathcal{M}_j\}
\end{equation}

\textbf{Step 2: Density Peak Estimation and Maximum Height Filtering.} Kernel Density Estimation (KDE) is employed to model the $z$-axis distribution of edge points, thereby facilitating the elimination of outliers and extraction of prominent edge features. Given the set $\{z_i\}_{i=1}^N$, which represents the heights of $N$ edge points obtained from the mask denoted by ${\widetilde{m}}_j$, the probability density function is approximated by
\begin{equation}
\label{eq22}
\hat{f}\left(z\right)=\frac{1}{Nh}\sum_{i=1}^{N}{K(\frac{z-z_i}{h})}
\end{equation}

\noindent where $K\left(\cdot\right)$ denotes the Gaussian kernel function and $h$ is the bandwidth. The density peak is defined as $z^\ast = {\arg\max}_{z} \hat{f}(z)$. All points within the height range $\left|z_i-z^\ast\right|\le\delta$ are retained, resulting in filtered 3D and 2D point sets $\mathcal{P}_j^{\text{KDE}}\subset\mathcal{P}_j$ and $\mathcal{M}_j^{\text{KDE}}\subset\mathcal{M}_j$. 
Subsequently, the maximum height along the $z$-axis in $\mathcal{P}_j^{\text{KDE}}$ is identified as $z_{max}$. Finally, points with heights satisfying $z_i{\geq z}_{max}-\eta$ are further selected, yielding the final point sets $\mathcal{P}_j^{\text{peak}}\subset\mathcal{P}_j^{\text{KDE}}$ and $\mathcal{M}_j^{\text{peak}}\subset\mathcal{M}_j^{\text{KDE}}$.

\textbf{Step 3: Centrally Symmetric Point Pairs as Affordance Targets.} Further integration of object position and structural information is achieved by identifying centrally symmetric point pairs, which are used to derive the refined affordance targets, denoted as $\mathcal{C}_i^{\text{init}}$. 
Given the course anchor $\bm{c}_j \in C_{\text{anchor}}$, we enumerate all point pairs $\bm{u}_a,\bm{u}_b\in\mathcal{M}_j^{\text{peak}}$ and identify the pair whose midpoint is closest to $\bm{c}_j$, defined as
\begin{equation}
\label{eq23}
(\bm{u}_a^\ast, \bm{u}_b^\ast) = 
\mathop{\arg\min}\limits_{\bm{u}_a, \bm{u}_b \in \mathcal{M}_j^{\text{peak}}}
\left\| \frac{\bm{u}_a + \bm{u}_b}{2} - \bm{c}_j \right\|_2
\end{equation}

\noindent where $\left(\bm{u}_a^\ast,\bm{u}_b^\ast\right)$ denotes the filtered point pair, with 3D coordinates $\bm{p}_a,\bm{p}_b\in\mathcal{P}_j^{\text{peak}}$. Finally, the refined affordance target $\mathcal{C}_i^{\text{init}}$ is expressed as
\begin{equation}
\label{eq24}
\mathcal{C}_i^{\text{init}}=\frac{1}{2}\left(\bm{p}_a+\bm{p}_b\right)
\end{equation}

\subsection{Real-Time Close-Loop Control Strategy}

To solve the constrained optimization problem in Equation \eqref{eq2} with real-time performance, the Model Predictive Path Integral (MPPI) control algorithm is employed on the Isaac Gym simulation platform. MPPI is a stochastic optimization method designed for robot control in dynamic and uncertain environments. Essentially, the algorithm samples multiple candidate control sequences from the control input space, evaluates the cost functions based on simulated state trajectories, and updates the control policy through importance-weighted averaging. The GPU-accelerated parallel simulation enabled by Isaac Gym significantly enhances sampling efficiency, thereby supporting real-time execution in dynamic environments. For a detailed explanation of MPPI, please refer to \cite{williams2018mppi}. In this work, we derive the specific formulation for solving the constrained optimization problem presented in Equation \eqref{eq2}.

To begin with, the current joint state of the robot is denoted as $\bm{x}(t) = \left[ {\begin{array}{*{20}{c}}
\bm \theta(t)  \\
\dot{\bm \theta}(t) 
\end{array}} \right] \in \mathbb{R}^{2n}$, comprising joint positions and velocities. With the velocity control input $\bm{v}(t)\in \mathbb{R}^n$, where $n$ represents the degrees of freedom, the nonlinear state transition function can be formulated as
\begin{equation}
\label{eq25}
\bm{x}(t+1)=S(\bm{x}(t),\bm{v}(t)),\ \ \bm{v}(t)\sim \mathcal{N}\left(\mu(t),\Sigma\right)
\end{equation}

\noindent where $\mathcal{N}\left(\cdot\right)$ denotes Gaussian sampling, $\mu(t)$ is the sampling mean, $\Sigma$ is the covariance matrix, and $S$ represents the forward kinematics and physical model. At each time step, MPPI samples $K$ distinct control sequences $\left\{\bm{V}_k\right\}_{k=1}^K$, where each sequence consists of velocity inputs over a time horizon $T$, expressed as $\bm{V}_k=[\bm{v}_k(0),\bm{v}_k(1),\ldots,\bm{v}_k(T-1)]^\top$. Subsequently, each state trajectory is derived by combining the initial state $\bm{x}(0)$ with the forward model as
\begin{equation}
\label{eq26}
\mathcal{X}_k=[\bm{x}_k(0),\bm{x}_k(1),\ldots,\bm{x}_k(T)]^\top
\end{equation}

\noindent where $\bm{x}_k(t+1)=S(\bm{x}_k(t),\bm{v}_k(t))$. We further introduce the forward kinematics mapping $\varphi:\ \mathbb{R}^{2n}\rightarrow SE(3)$, which maps the joint space state to the end-effector pose, expressed as $\bm{T}_e^k(t)=\varphi(\bm{x}_k(t))$. At this point, $\bm{T}_e^k(t)\in SE(3)$ denotes the end-effector pose of the $k$-th trajectory at time $t$. For each trajectory $\mathcal{X}_k$, the total cost is the cumulative sum of the objective function $\mathcal{J}\left(\cdot\right)$ over each time step, defined as
\begin{equation}
\label{eq27}
C_k=\sum_{t=0}^{T-1}\mathcal{J}\left(\bm{T}_e^k(t), \mathcal{C}_i\right)
\end{equation}

For convenience in the analysis, we define the position and orientation errors as follows:
\begin{equation}
\label{eq28}
    \mathcal{D}_p=d_p\left(P(\bm{T}_e^k(t)), P(\mathcal{C}_i)\right)
\end{equation}
\begin{equation}
\label{eq29}
      \mathcal{D}_r=d_r\left(R(\bm{T}_e^k(t)), R(\mathcal{C}_i)\right)
\end{equation}

\noindent where $P(\cdot)$ and $R(\cdot)$ denote the position and quaternion components. The errors are calculated as $d_p(\bm{p}_1,\bm{p}_2)={\left \|\bm{p}_1-\bm{p}_2\right \|}_2$ and $d_r(\bm{q}_1,\bm{q}_2)=2\, \text{arccos}(\text{max}(-1,\text{min}(1,\langle \bm{q}_1,\bm{q}_2 \rangle)))$. Subsequently, the objective function is defined as
\begin{align}
\label{eq30}
\mathcal{J}\left(\bm{T}_e^k(t), \mathcal{C}_i\right) 
= \lambda_p \cdot \mathcal{D}_p + \lambda_r \cdot\mathcal{D}_r + \lambda_c \cdot\mathcal{D}_c
\end{align}

\noindent where $\lambda_p$ and $\lambda_r$ are the weights for position and orientation costs, $\lambda_c$ denote the weight for the collision cost $\mathcal{D}_c$, which is formulated based on the ESDF representation, as described in \cite{huang2024rekep}. In this work, the objective function is subject to preconditions and postconditions as specified in Equation \eqref{eq2}, the precondition typically ensures that the end-effector is within the designated region prior to the execution stage, while the postcondition assesses whether the end-effector has successfully reached the desired pose, defined as
\begin{equation}
\label{eq31}
\epsilon_{\text{pre}}\left(\bm{T}_e^k(t),\ \mathcal{C}_i\right)=\mathcal{D}_p(k, t, i)
\end{equation}
\begin{equation}
\label{eq32}
\epsilon_{\text{post}}\left(\bm{T}_e^k(t),\mathcal{C}_i\right)=\mathcal{D}_p(k, t, i)+\mathcal{D}_r(k, t, i)
\end{equation}

After evaluating the total cost $C_k$, MPPI assigns weights to each sampled trajectory as follows:
\begin{equation}
\label{eq33}
\omega_k=\frac{1}{\eta}\exp\left(-\frac{1}{\beta}\left(C_k-\rho\right)\right),\ \ \sum_{k=1}^{K}{\omega_k=1}
\end{equation}

\noindent where $\beta$ is the temperature parameter governing the variance of sample weights, and $\rho=\min_k{C_k}$ is the stability offset. The approximate optimal control sequence $\bm{U}^\ast=\left[\bm{u}_0^\ast,\bm{u}_1^\ast,\ldots,\bm{u}_{T-1}^\ast\right]$ is expressed as
\begin{equation}
\label{eq34}
\bm{U}^\ast=\sum_{k=1}^{K}{\omega_k\cdot \bm{V}_k}
\end{equation}

Finally, the first control input $\bm{u}_0^\ast$ of the sequence $\bm{U}^\ast$ is applied as the velocity command to the system, and the optimization procedure is conducted iteratively.

\begin{figure}[!t]
\centering
\includegraphics[width=\columnwidth]{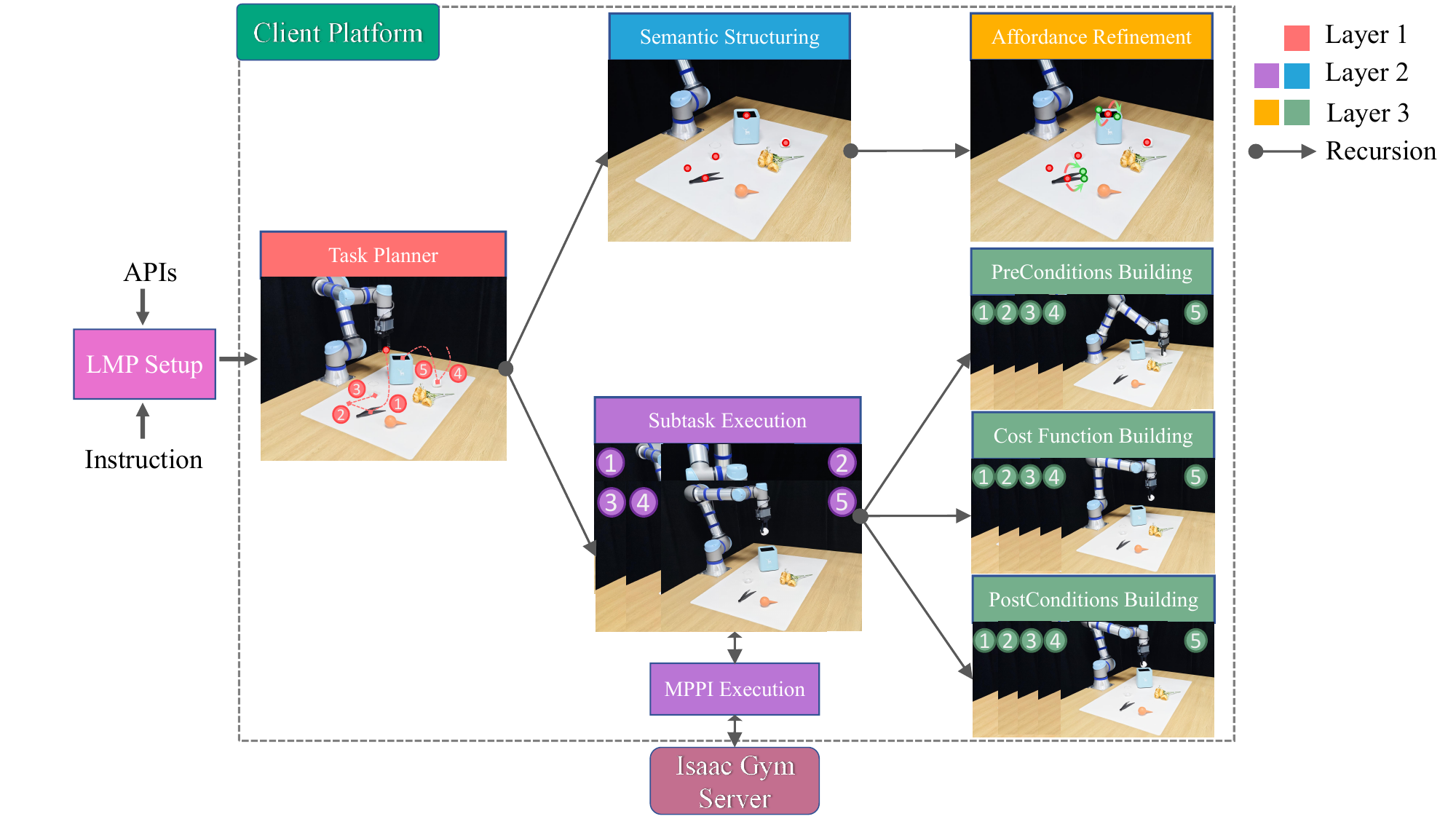}
\caption{MLLM-driven automated modeling and communication framework for TAMP.}
\label{fig_4}
\end{figure}

\subsection{MLLM-driven Automated Modeling of TAMP}

As depicted in Fig.~\ref{fig_4}, the Language Model Program (LMP) framework is built on a hierarchical recursive architecture \cite{liang2023cap}, enabling structured invocation and multi-level prompt nesting. Within this framework, we encapsulated seven LMP modules, including Task Planner, Semantic Structuring, Affordance Refinement, Subtask Execution, Preconditions Building, Cost Function Building, and PostConditions Building. Each module implements an independent prompt structure and interfaces with system components through standardized APIs, enabling automated modeling of task and motion planning.

During the LMP Setup, dependency relationships among LMPs are established and mappings to APIs are configured driven by the natural language instructions. Subsequently, Task Planner performs global scheduling and intent parsing, collaborating with Semantic Structuring and Subtask Execution to model affordance targets and execute tasks. Finally, Affordance Refinement further refines these targets through geometric and positional flows with recursive feedback.

In the Subtask Execution phase, tasks are decomposed into constrained action execution units. Specifically, initiation conditions, optimization objectives, and termination criteria are formulated respectively by Preconditions Building, Cost Function Building, and Postconditions Building. Subsequently, each action unit activates the MPPI Execution, establishing the communication loop between the client and Isaac Gym server. The client transmits real-time joint states and perception-fused cost functions, while the server performs GPU-accelerated MPPI sampling and optimization to synthesize continuous joint-space velocity commands, which are transmitted back to the client, enabling stepwise closed-loop control.

\begin{figure*}[!t]
\centering
\includegraphics[width=0.85\textwidth]{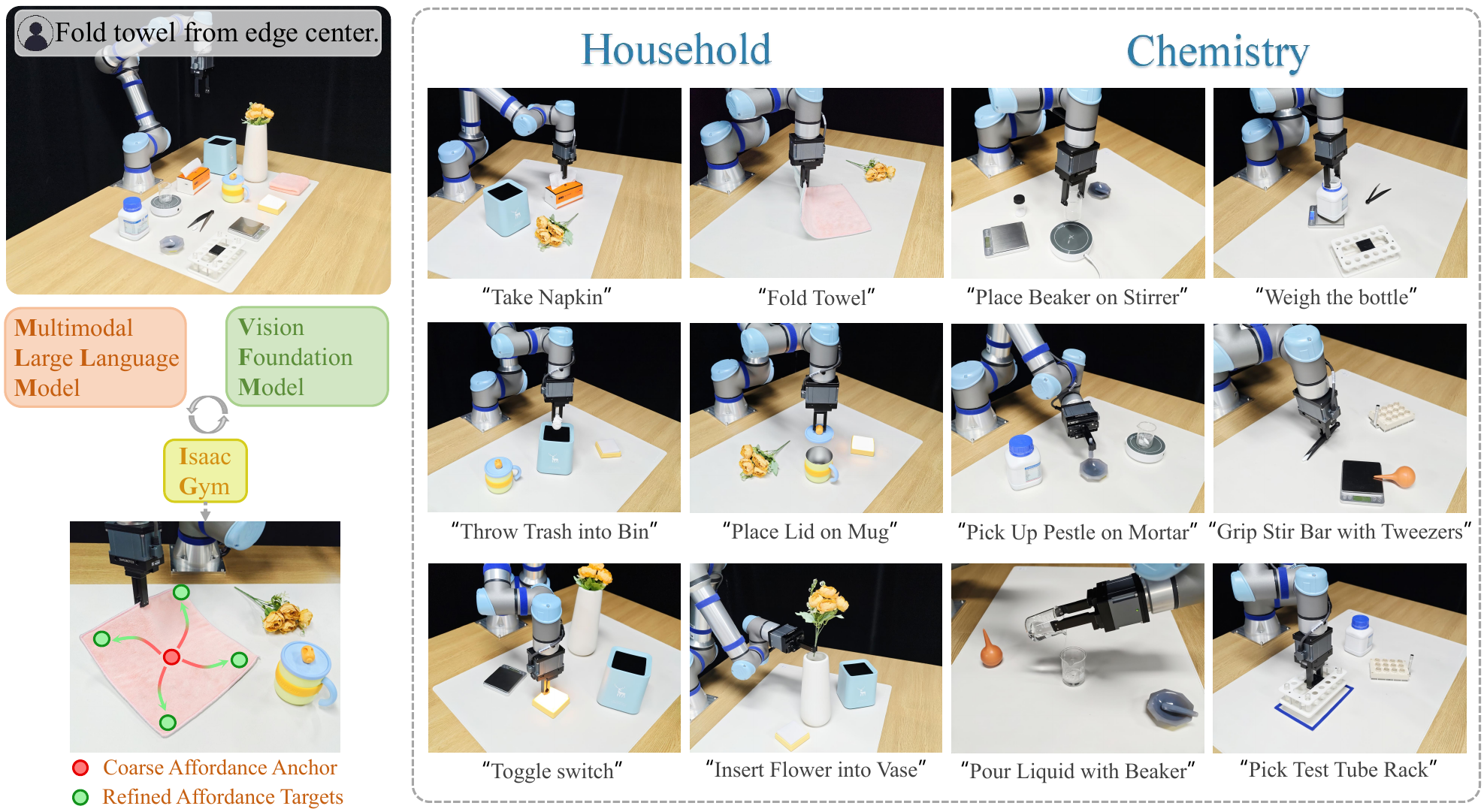}
\caption{ReSemAct is a unified robotic manipulation framework for semantically diverse environments. It leverages synergistic reasoning between MLLMs and VFMs, introducing Semantic Structuring and Affordance Refinement (SSAR) to enable fine-grained robotic manipulation.}
\label{fig_5}
\end{figure*}

In the course of task execution, preconditions and postconditions are monitored. Upon violation of the precondition, the system backtracks and restarts the previous subtask, whereas satisfaction of the postcondition triggers progression to the next subtask. Ultimately, ReSemAct establishes a closed-loop TAMP framework centered on semantic understanding, guided by hierarchical recursion, and grounded in constraint feedback.

\section{Experiment}
\label{Section IV}
In this section, we first detail the experimental setup. Subsequently, we evaluate the cross-platform generalization of ReSemAct in simulated environments. Then, we demonstrate the manipulation generalization in the semantically diverse real-world  environments. Furthermore, we conduct ablation experiment to evaluate the effectiveness of SSAR strategy. Finally, we analyze the core mechanisms of ReSemAct and identify the key factors contributing to execution failures.

\subsection{Experimental Setup}

\begin{figure}[!t]
\centering
\includegraphics[width=\columnwidth]{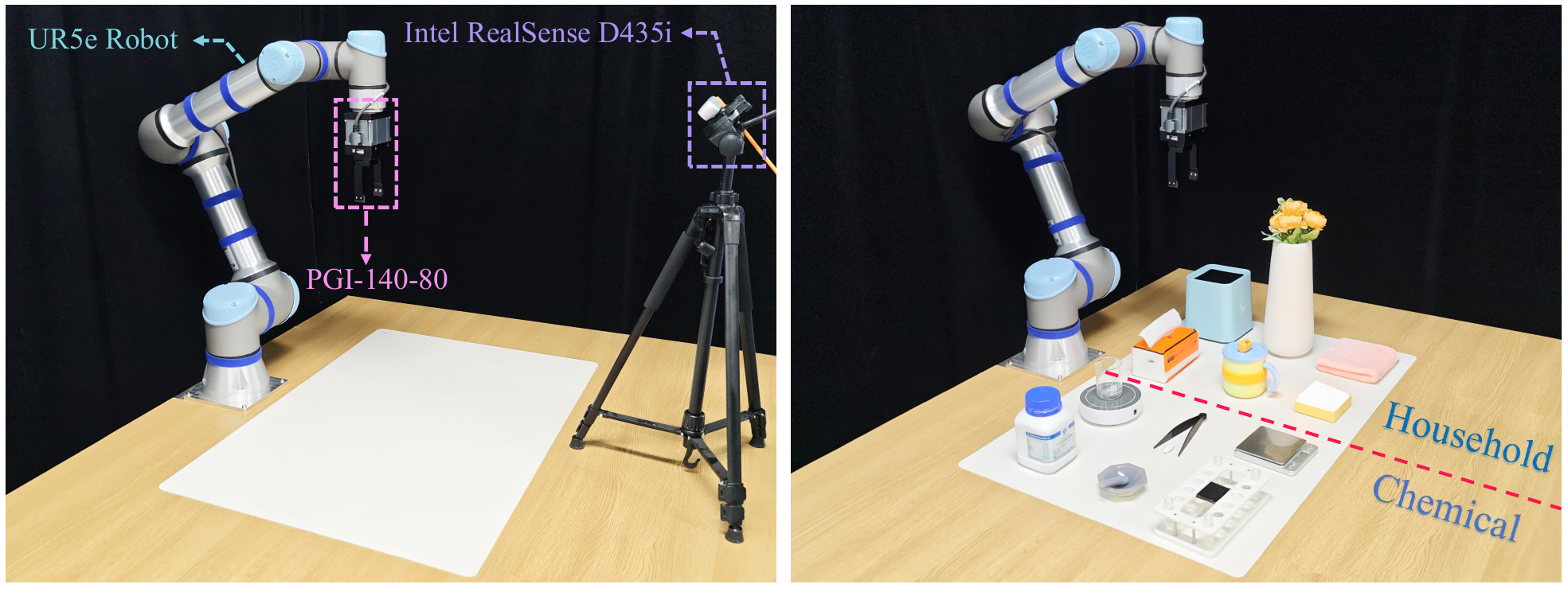}
\caption{Hardware Platform Configuration.}
\label{fig_6}
\end{figure}

\textbf{Hardware Configuration.} As shown in Fig.~\ref{fig_6}, our real-world experimental platform is built around the 6-DOF Universal Robots UR5e robotic arm, equipped with a PGI-140-80 two-finger parallel gripper and an Intel RealSense D435i RGB-D camera. For simulation, we employ the Omnigibson \cite{li2023behavior1k} platform developed by Stanford University, and utilize both the UR5e and the 7-DOF Franka Panda collaborative arms.

\textbf{Tasks and Metrics.} As shown in Fig.~\ref{fig_5}, we designed 12 real-world tasks (6 chemical lab scenarios and 6 household scenarios) to evaluate generalization and stability under static conditions and dynamic disturbances. In simulation, we perform 10 tasks (6 chemical lab scenarios and 4 household scenarios) to evaluate generalization across different robotic platforms under static conditions. Furthermore, we designed three simulated and four real-world experiments to assess the effectiveness in semantic grounding. All tasks were evaluated over 10 trials, and success rate was reported.

\textbf{Baselines.} We compare against three representative frameworks: VoxPoser \cite{huang2023voxposer}, ReKep \cite{huang2024rekep} and CoPa \cite{huang2024copa}. VoxPoser integrates large language and vision models to generate 3D value maps interpretable by robots, enabling closed-loop trajectory synthesis without predefined motion primitives. It leverages GPT-4 and a visual frontend for semantic grounding of task objects, supporting flexible manipulation across diverse tasks.  ReKep fuses VLMs and MLLMs to infer semantic keypoint constraints for structured tasks and solves end-effector trajectories in real time via hierarchical optimization. CoPa leverages foundation vision-language models to propose a two-stage task execution process, performing task-oriented grasping via coarse-to-fine visual grounding in the first stage, and identifying task-relevant spatial and geometric constraints for robot planning in the second stage. All three frameworks deeply integrate large language and vision models, employing semantic grounding and control optimization to achieve efficient and robust robotic manipulation.

\textbf{VFM and MLLM.} We utilize GPT-4o \cite{achiam2023gpt-4} from OpenAI as a callable MLLM and combine it with the visual foundation model FastSAM \cite{zhao2023fastsam} to enable high-efficiency multimodal reasoning (prompt design is available at \href{https://resemact.github.io}{ReSemAct.github.io}. During the dynamic disturbance phase, we incorporate the TAPNext model from Google DeepMind \cite{zholus2025tapnext} together with real-time RGB frame input to achieve online point tracking at approximately 20 Hz. This facilitates real-time inference of evolving affordance targets in dynamic environments, thereby enhancing the stability and robustness of task execution.

\begin{table*}[!t]
\centering
\caption{Success Rates across Static Tasks in Simulation}
\label{tab1}
\renewcommand{\arraystretch}{1.1}
\begin{tabular}{llccc}
\toprule
\textbf{Household Scenario} & \textbf{Robot Platform} &
\textbf{Voxposer} & \textbf{Rekep} & \textbf{ReSemAct (Ours)} \\
\midrule
Toggle switch & UR5e & 0/10 & 3/10 & \textbf{8/10} \\
Fold Towel & Franka Panda & 0/10 & 3/10 & \textbf{5/10} \\
Take Napkin & Franka Panda & \textbf{8/10} & 6/10 & 7/10 \\
Cut Cake with Knife & Franka Panda & 0/10 & 5/10 & \textbf{6/10} \\
\midrule
\textbf{Total} &  & 20.0\% & 42.5\% & \textbf{65.0\%} \\
\midrule
\textbf{Chemical Lab Scenario} & \textbf{Robot Platform} &
\textbf{Voxposer} & \textbf{Rekep} & \textbf{ReSemAct (Ours)} \\
\midrule
Pour reagent into dish & UR5e & 7/10 & 0/10 & \textbf{7/10} \\
Weigh the Reagent Bottle & UR5e & \textbf{8/10} & 1/10 & \textbf{8/10} \\
Pick Up Pestle on Mortar & UR5e & 0/10 & 3/10 & \textbf{5/10} \\
Place Reagent Bottle on Stirrer & UR5e & 7/10 & 2/10 & \textbf{8/10} \\
Place Funnel on Iron Ring & Franka Panda & 0/10 & 0/10 & \textbf{4/10} \\
Grip Stir Bar with Tweezers & Franka Panda & 0/10 & 0/10 & \textbf{5/10} \\
\midrule
\textbf{Total} &  & 36.6\% & 10.0\% & \textbf{61.6\%} \\
\bottomrule
\end{tabular}
\end{table*}

\begin{table}[!t]
\centering
\caption{Semantic Grounding Success Rates in Simulation}
\label{tab2}
\renewcommand{\arraystretch}{1}
\begin{tabular}{llcc}
\toprule
\textbf{Task} & \textbf{Scale} & \textbf{ReKep} & \textbf{ReSemAct (Ours)} \\
\midrule
\multirow{3}{*}{Insert Flower into Vase}
  & 150\% & 3/10 & \textbf{7/10} \\
  & 70\%  & 1/10 & \textbf{6/10} \\
  & 40\%  & 0/10 & \textbf{4/10} \\
\midrule
\multirow{3}{*}{Insert Bread into Toaster}
  & 150\% & 4/10 & \textbf{7/10} \\
  & 70\%  & 0/10 & \textbf{8/10} \\
  & 40\%  & 0/10 & \textbf{8/10} \\
\midrule
\multirow{3}{*}{Pick Up Pestle on Mortar}
  & 150\% & \textbf{7/10} & \textbf{7/10} \\
  & 70\%  & 6/10 & \textbf{7/10} \\
  & 40\%  & 1/10 & \textbf{8/10} \\
\bottomrule
\end{tabular}
\end{table}

\subsection{Evaluation of ReSemAct in Simulated Environments}

As summarized in Table~\ref{tab1}, we systematically evaluate the generalization of ReSemAct in simulation across different robotic platforms and semantically diverse environments, comprising 4 household and 6 chemical lab tasks under static conditions. For the Toggle Switch task, the density peak in positional refinement Flow is replaced by the maximum height to enable more direct extraction of symmetric point pairs. The results demonstrate that ReSemAct exhibits strong task adaptability and cross-platform generalization, achieving execution success rates of 65\% in household and 61.6\% in chemical lab scenarios, respectively—consistently outperforming the baselines. This performance is attributed to the proposed SSAR. Specifically, VoxPoser achieves competent performance in tasks where the 3D center point of the object mask spatially aligns with a feasible grasp point (e.g., napkin, reagent, and stirrer). However, the success rate drops markedly when such alignment is absent. Although Rekep consistently generates semantically meaningful affordance targets across tasks, its effectiveness is limited by the mismatch between predicted and task-relevant affordance targets. Notably, since depth information for transparent containers is available in Omnigibson, these objects are modeled as opaque (e.g., plastic bottles) in simulation. Despite differences from real-world visual characteristics, this scheme preserves consistency in target generation, ensuring rigor in evaluation.

\begin{figure}[!t]
\centering
\begin{minipage}{\columnwidth}
  \centering
  \includegraphics[width=\columnwidth]{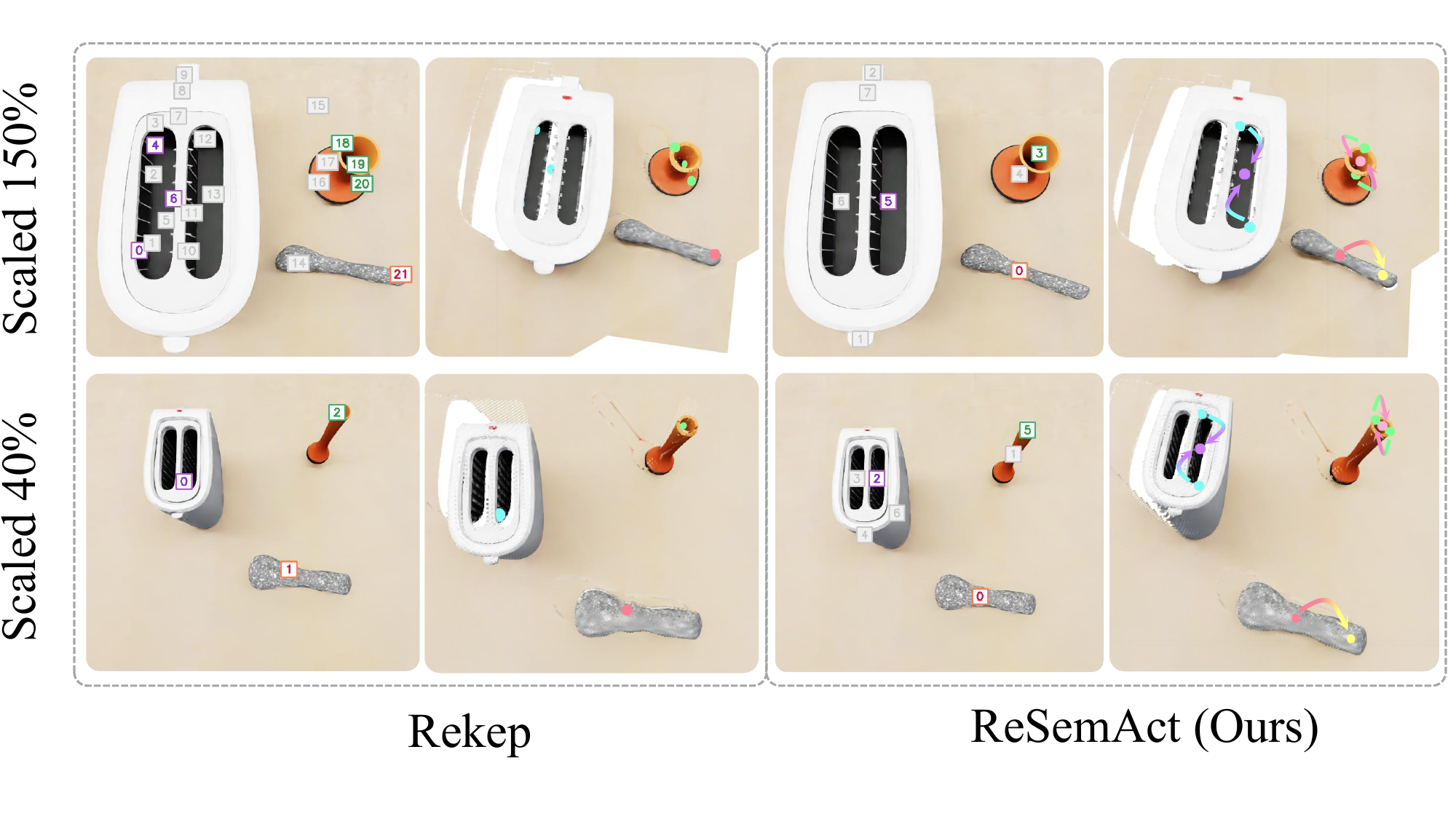}
  \caption{Comparison between Rekep and \textbf{ReSemAct (Ours)} in semantic grounding. Each method has two columns presenting RGB and point cloud visualizations, and two rows for 150\% and 40\% object scales. The numerical labels in purple, green, and red indicate the initial 2D constraints extracted by MLLM on the toaster, vase, and pestle, respectively, while the colorful arrows indicate the geometric and positional refinement.}
  \label{fig_7}
\end{minipage}
\end{figure}

For further analysis, we focus on the affordance targets adopted by ReKep for open-structured containers and dual-ended semantic objects. As shown in Fig.~\ref{fig_7}, the affordance targets of such objects are typically extracted by the MLLM, either computing the centroid of multiple opening-edge points (e.g., the vase and toaster) or identifying functionally distinct endpoints (e.g., the narrower end of the pestle). However, ReKep is sensitive to affordance targets spacing and considerably dependent on object scale: scale increases tend to produce redundant targets that disrupt accurate recognition, while scale decreases often cause targets merging, leading to the loss of affordance targets. By comparison, ReSemAct demonstrates significantly greater stability under scale variation, exhibiting enhanced robustness against scale perturbations. This advantage is substantiated in Table~\ref{tab2}, where ReSemAct consistently outperforms baseline methods in semantic grounding success rates on the Vase, Toaster and Pestle scaled to 150\%, 70\%, and 40\%. These results indicate that the method effectively alleviates the impact of scale variation in open-structured container and dual-ended semantic object manipulation, thereby enhancing the fine-grained robot manipulation.

\begin{table*}[t]
\centering
\caption{Success Rates in Real-world Environments}
\label{tab3}
\renewcommand{\arraystretch}{1.1}
\begin{tabular}{lcccccc}
\toprule
\textbf{Task Scenarios} &
\multicolumn{2}{c}{\textbf{Voxposer}} &
\multicolumn{2}{c}{\textbf{Rekep}} &
\multicolumn{2}{c}{\textbf{ReSemAct (Ours)}} \\
\midrule
\textbf{Household Scenario} &
\textbf{Static} & \textbf{Dist.} &
\textbf{Static} & \textbf{Dist.} &
\textbf{Static} & \textbf{Dist.} \\
\midrule
Take Napkin             & 3/10 & 2/10 & 5/10 & 4/10 & \textbf{7/10} & \textbf{5/10} \\
Fold Towel              & 0/10 & 0/10 & 2/10 & 1/10 & \textbf{5/10} & \textbf{3/10} \\
Place Lid on Mug        & 0/10 & 0/10 & 3/10 & 1/10 & \textbf{5/10} & \textbf{4/10} \\
Toggle switch           & 0/10 & 0/10 & 1/10 & 0/10 & \textbf{4/10} & \textbf{3/10} \\
Insert Flower into Vase & 0/10 & 0/10 & 2/10 & 1/10 & \textbf{6/10} & \textbf{4/10} \\
Throw Trash into Bin    & 7/10 & 6/10 & 6/10 & 5/10 & \textbf{8/10} & \textbf{7/10} \\
\midrule
\textbf{Total}          & 16.6\% & 13.3\% & 31.6\% & 20.0\% & \textbf{58.3\%} & \textbf{43.3\%} \\
\midrule
\textbf{Chemical Lab Scenario} & & & & & & \\
\midrule
Place Beaker on Stirrer     & 0/10 & 0/10 & 1/10 & 0/10 & \textbf{6/10} & \textbf{5/10} \\
Pour Liquid with Beaker     & 0/10 & 0/10 & 1/10 & 0/10 & \textbf{7/10} & \textbf{6/10} \\
Pick Test Tube Rack         & 1/10 & 0/10 & 1/10 & 0/10 & \textbf{6/10} & \textbf{5/10} \\
Pick Up Pestle on Mortar    & 0/10 & 0/10 & 1/10 & 0/10 & \textbf{5/10} & \textbf{4/10} \\
Grip Stir Bar with Tweezers & 0/10 & 0/10 & 0/10 & 0/10 & \textbf{4/10} & \textbf{2/10} \\
Weigh the Bottle            & 7/10 & 5/10 & 5/10 & 4/10 & \textbf{8/10} & \textbf{6/10} \\
\midrule
\textbf{Total}              & 13.3\% & 8.3\% & 15.0\% & 6.6\% & \textbf{60.0\%} & \textbf{46.6\%} \\
\bottomrule
\end{tabular}
\end{table*}

\begin{table}[!t]
\centering
\caption{Semantic Grounding Success Rates in Real World}
\label{tab4}
\renewcommand{\arraystretch}{1}
\begin{tabular}{lcc}
\toprule
\textbf{Task} & \textbf{CoPa} & \textbf{ReSemAct (Ours)} \\
\midrule
Fold Towel & 0/10 & \textbf{6/10} \\
Toggle Switch & 1/10 & \textbf{5/10} \\
Pick Up Pestle on Mortar & 0/10 & \textbf{7/10} \\
Grip Stir Bar with Tweezers & 0/10 & \textbf{5/10} \\
\midrule
\textbf{Total} & 2.5\% & \textbf{57.5\%} \\
\bottomrule
\end{tabular}
\end{table}

\subsection{Evaluation of ReSemAct in real-world Environments}

In this section, we systematically evaluate the generalization of ReSemAct in the real world through semantically diverse manipulation tasks, including 6 household and 6 chemical lab scenarios. As shown in Table~\ref{tab3}, “Static” denotes environments without external disturbances, whereas “Dist.” refers to dynamic disturbances from manually displacing objects. Compared with baselines, ReSemAct achieves significantly superior performance, with 58.3\% (Static) and 43.3\% (Dist.) in the household scenario, and 60\% (Static) and 46.6\% (Dist.) in the chemical lab scenario, demonstrating enhanced generalization and robustness against dynamic disturbances. A comprehensive analysis is presented below.

Specifically, household scenarios are characterized by rich semantic structures and modest manipulation precision, which enable baselines to retain considerable generalization in semantic grounding. Voxposer is robust in tasks where the 3D center points of object masks (e.g., napkins, trash, bins) are directly graspable, consistent with its simulation performance. For other tasks, failure occurs when the localization results do not meet the grasping requirements. Additionally, Rekep effectively extracts rich semantic affordance targets for MLLM interpretation. However, complex semantic structures and fine-grained targets (e.g., locating towel corners, detecting raised switch edges) surpass its interpretative capabilities. In contrast, chemical lab scenarios are typified by sparse semantic structure and prevalence of transparent materials, which degrade depth perception and necessitate greater spatial precision. These conditions further constrain baseline generalization (e.g., localizing beaker center, identifying pestle’s narrow tip), causing a continued decline in success rates. Building on these insights, ReSemAct introduces a Semantic Stucturing and Affordance Refinement Strategy that integrates fine-grained semantic grounding, achieving 1.7\% (Static) and 3.3\% (Dist.) improvement in task success rates, maintaining strong robustness in semantically diverse environments.

\begin{figure}[!t]
\centering
\begin{minipage}{\columnwidth}
  \centering
  \includegraphics[width=\columnwidth]{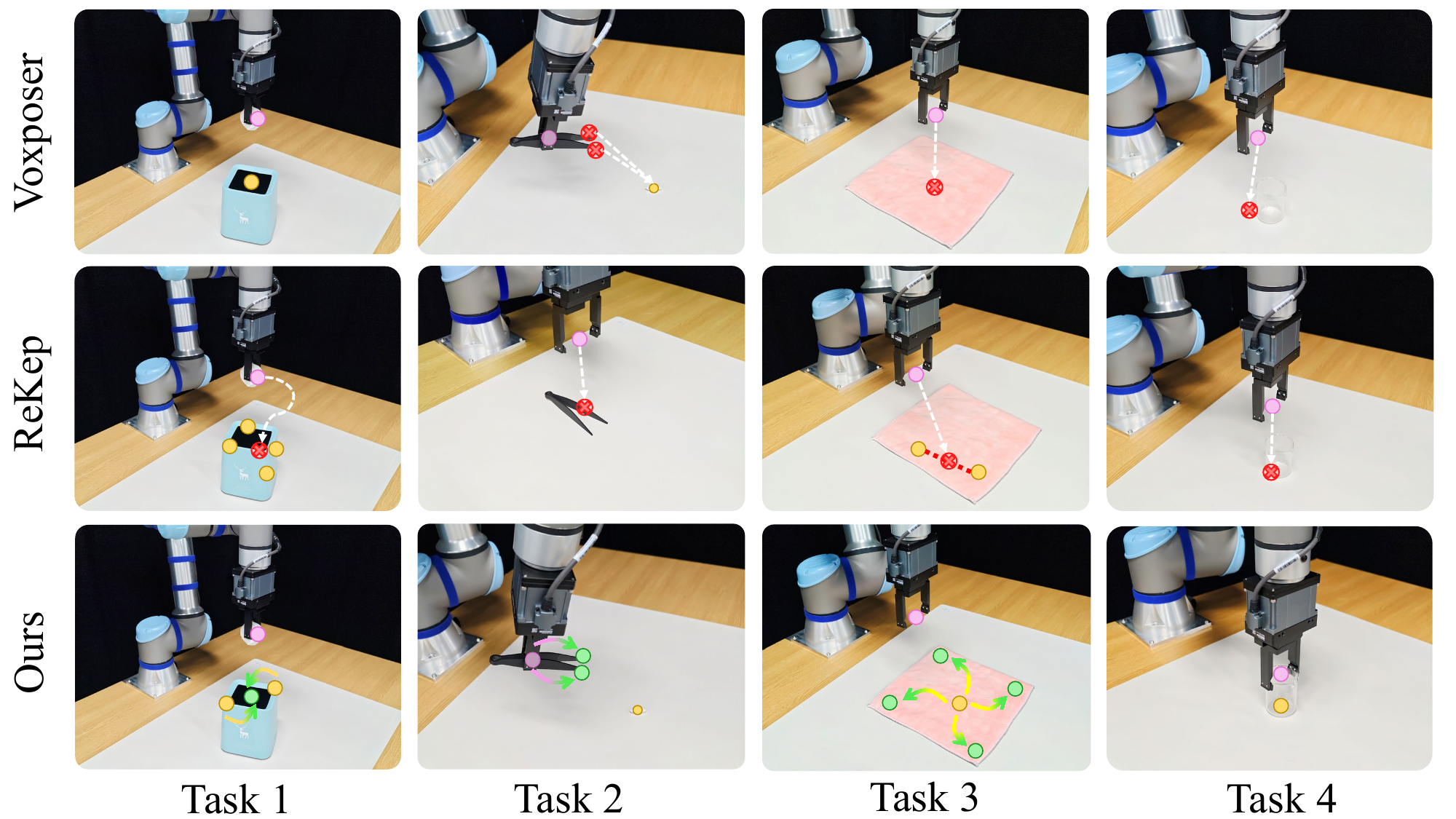}
  \caption{Comparison between Baselines and \textbf{ReSemAct (Ours)} in semantic grounding and closed-Loop execution. Tasks 1 to 4 correspond to the execution for the bin, tweezer, towel, and beaker. The pink marks indicate end-effector positions or grasped targets, yellow marks denote extracted course affordance anchor, green marks represent refined affordance targets, red marks highlight non-executable targets, and colorful arrows show target refinement.}
  \label{fig_8}
\end{minipage}
\end{figure}

To qualitatively analyze the advantages of ReSemAct, we focus on its performance in semantic grounding and closed-loop execution efficiency, as shown in Fig.~\ref{fig_8}. It can be observed that the affordance targets from VoxPoser mainly concentrate around the object center, lacking fine-grained perception. Additionally, the absence of point clouds for transparent objects in chemical scenarios (e.g., beaker in Task 4) presents a significant challenge to accurate localization. Furthermore, the 5Hz 3D value map update rate restricts the responsiveness of closed-loop trajectory planning. Similarly, while ReKep constructs semantically directed targets, it is limited in the accuracy (e.g., the center of the beaker, bin, and tweezer) and refinement (e.g., the corners of the towel, the tips of the tweezer) of affordance targets. Its performance depends on the effectiveness of affordance generation and high-level reasoning by MLLM. Moreover, closed-loop optimization at 10Hz is constrained by the efficiency of inverse kinematics solving, leading to unstable real-time responsiveness. 

\begin{figure}[!t]
\centering
\begin{minipage}{\columnwidth}
  \centering
  \includegraphics[width=\columnwidth]{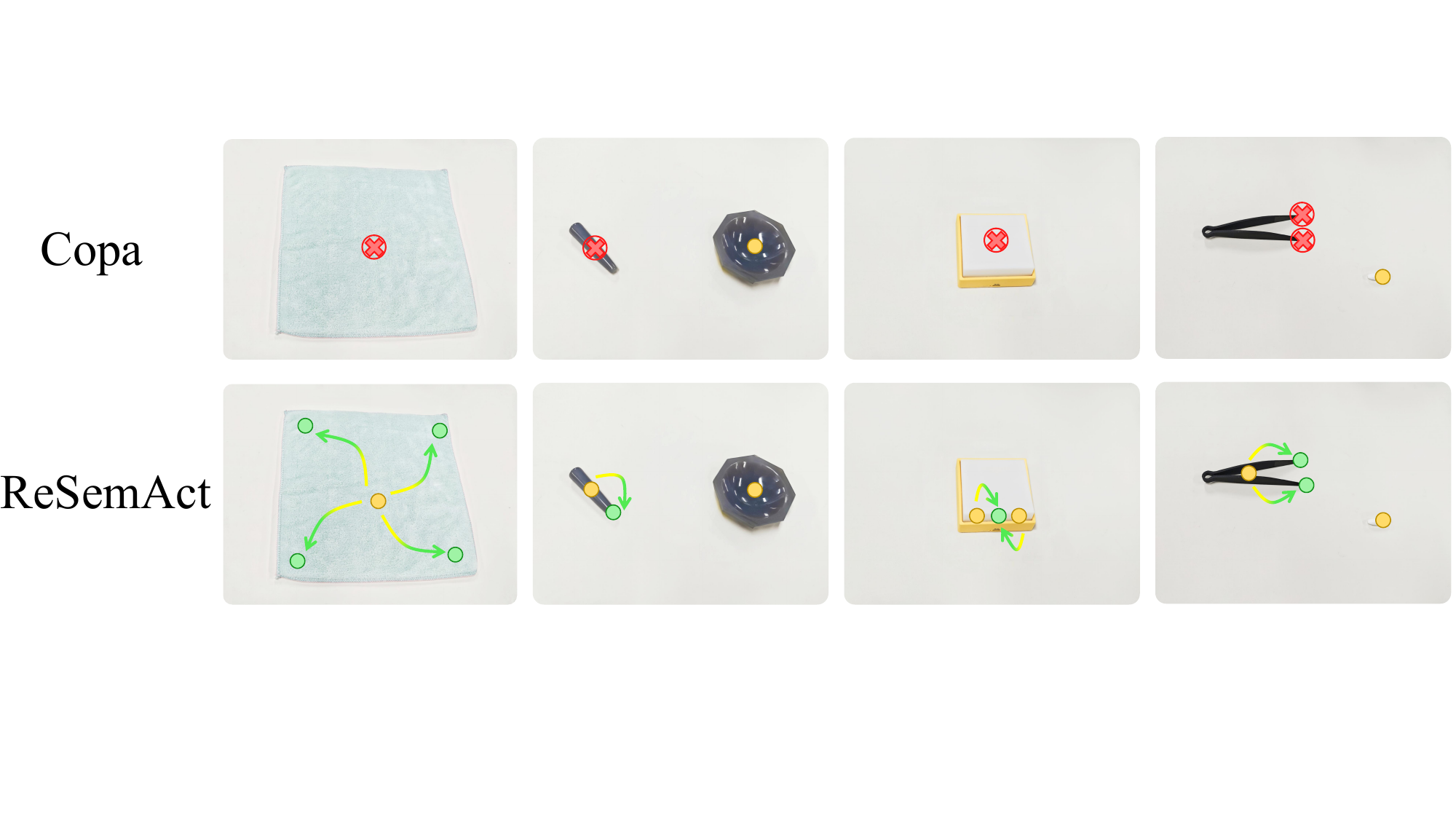}
  \caption{Semantic grounding comparision between Copa and \textbf{ReSemAct}.}
  \label{fig_9}
\end{minipage}
\end{figure}

Furthermore, we compare with CoPa, which also adopts fine-grained robot manipulation. As shown in Table~\ref{tab4}, we evaluate four representative tasks from Table~\ref{tab3} to examine whether the semantic grounding is sufficient for successful task execution. Experimental results show that ReSemAct achieves a 57.5\% success rate, significantly outperforming CoPa. A detailed qualitative analysis in Fig.~\ref{fig_9} shows that, while CoPa performs hierarchical grounding, its fine-grained localization precision is limited to the part level, making it insufficient for tasks requiring finer spatial reasoning.

\subsection{Ablation Experiment}
\begin{figure}[!t]
\centering
\begin{minipage}{\columnwidth}
  \centering
  \includegraphics[width=0.9\columnwidth]{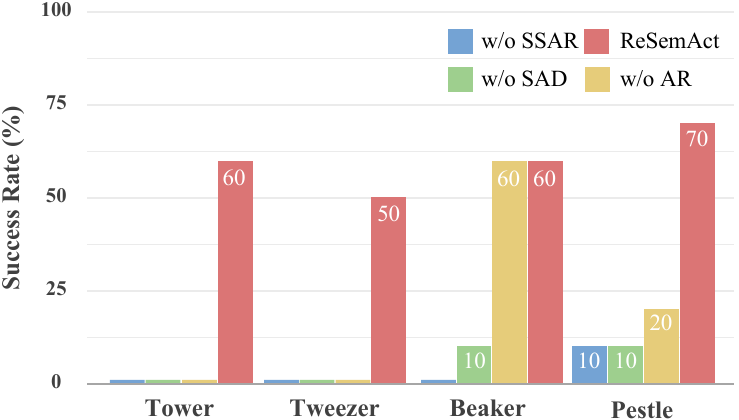}
  \caption{ReSemAct Ablation experiments on Four Tasks.}
  \label{fig_10}
\end{minipage}
\end{figure}

In this section, we conduct an ablation Experiment to evaluate the effectiveness of Semantic Structuring and Affordance Refinement (SSAR), as shown in Fig.~\ref{fig_10}. We perform semantic grounding on four representative tasks involving the tower, tweezer, beaker, and pestle. Specifically, w/o SSAR denotes a variant that directly relies on GPT-4o to predict affordance targets under language and visual prompts. w/o Semantic Affordance Description (SAD) replaces the affordance region with that of all objects, disrupting the semantic structure while maintaining normal operation. w/o AR removes the affordance refinement flow for reasoning.

The experimental results show that directly relying on MLLMs to predict affordance targets makes the tasks nearly infeasible. Similarly, once the semantic structure in SAD is disrupted, the system is unable to perform further fine-grained refinement. In addition, removing the affordance refinement has a limited impact on the Beaker task, as its affordance target largely coincides with the coarse-grained anchor. ReSemAct achieves the highest success rate across all tasks, demonstrating that SSAR plays a crucial role in fine-grained semantic grounding and significantly improves task success rates.

\section{Discussion}
\label{Section V}
Given challenges to achieve fine-grained semantic grounding and maintaining real-time closed-loop manipulation, ReSemAct achieves better performance via two strategies: (1) \textbf{Semantic Structuring and Affordance Refinement.} A unified semantic affordance description is introduced to guide the affordance refinement, enabling fine-grained semantic grounding and robotic manipulation. (2) \textbf{MLLM-driven Automated TAMP Modeling.} ReSemAct autonomously achieves fine-grained semantic grounding via SSAR under multimodal reasoning and generates cost functions for the constrained optimizer. Additionally, it decomposes multi-stage tasks and generates preconditions and postconditions, enabling temporal management and dynamic backtracking during execution.

Ultimately, the stable operation of ReSemAct requires effective coordination across functional modules. To systematically identify failure causes, we conduct independent validation for each sub-module by calculating the failure rate based on Table~\ref{tab3}. Specifically, since the VFM module lacks semantic guidance, it may fail to provide task-relevant semantic grounding in 7\% of cases, thereby generating irrelevant candidates. Reasoning failures in semantic structuring and affordance refinement, constituting 16\% and 28\% of failures respectively, arise from insufficient cross-modal semantic grounding and limited spatial reasoning. The Constrained Optimization module often suffers from slow convergence under significant pose variation in 11\% of occurrences, due to multi-objective complexity. The TAPNext module exhibits reduced tracking accuracy under occlusion or inaccurate point cloud mappings in 38\% of situations, thereby breaking the control loop. 
To tackle these challenges, future work will explore the integration of MLLM-based self-correction strategies, global optimization algorithms, and multi-view fusion methods to further enhance the robustness and adaptability of the framework.

\section{Conclusion}
\label{Section VI}
In this work, we present ReSemAct, a unified robotic manipulation framework for semantically diverse environments. By leveraging synergistic reasoning between MLLMs and VFMs, we introduces Semantic Structuring and Affordance Refinement (SSAR) to achieve fine-grained robot manipulation. The refined targets are embedded into a automated TAMP framework and formulated as optimizable cost functions, enabling real-time closed-loop control. The proposed approach supports multi-stage task execution and condition-aware dynamic backtracking. Extensive experiments demonstrate that ReSemAct significantly outperforms existing methods across semantically diverse environments and multiple robotic platforms.

\bibliographystyle{IEEEtran}
\bibliography{reference}

\end{document}